\documentclass[a4paper, 10pt, conference]{IEEEtran}  
\IEEEoverridecommandlockouts
\usepackage[pdftex]{graphicx}
\def\BibTeX{{\rm B\kern-.05em{\sc i\kern-.025em b}\kern-.08em
    T\kern-.1667em\lower.7ex\hbox{E}\kern-.125emX}}
\usepackage{cite}
\usepackage{textcomp}
\usepackage{comment}

\usepackage{lmodern}

\usepackage{amsmath}
\usepackage[psamsfonts]{amssymb}
\usepackage{amsfonts}
\usepackage{braket}
\usepackage{latexsym}

\usepackage{multirow}

\IEEEpubid{\begin{minipage}{\textwidth}\ \\[70pt]
\centering\normalsize{\copyright 2021 IEEE}
\end{minipage}}

\begin{document}




\newcommand{\oomoji}[1]{{\mbox{\boldmath $#1$}}}
\newcommand{\tenti}{^{\rm T}}

\title{Generation of Gradient-Preserving Images allowing HOG Feature Extraction}

\author{\IEEEauthorblockN{1\textsuperscript{st} Masaki Kitayama}
\IEEEauthorblockA{\textit{Tokyo Metropolitan University}, \\ Tokyo, Japan \\
}
\and
\IEEEauthorblockN{2\textsuperscript{nd} Hitoshi Kiya}
\IEEEauthorblockA{\textit{Tokyo Metropolitan University}, \\ Tokyo, Japan \\
}
}

\maketitle

\begin{abstract}
In this paper, we propose a method for generating visually protected images, referred to as gradient-preserving images.
The protected images allow us to directly extract Histogram-of-Oriented-Gradients (HOG) features for privacy-preserving machine learning. In an experiment, HOG features extracted from gradient-preserving images are applied to a face recognition algorithm to demonstrate the effectiveness of the proposed method.
\end{abstract}

\section{Introduction}

The use of cloud environments and machine learning has greatly increased, so many studies
on privacy preserving machine learning with perceptually encrypted images have been reported\cite{sirichotedumrong2019privacy,sirichotedumrong2019pixel,maekawa2019privacy,kawamura2020privacy}.
However, most conventional methods cannot be applied to machine learning algorithms with Histogram-of-Oriented-Gradients (HOG)\cite{HOG} features, which are well-known to be effective in many applications.

Accordingly, in this paper, we propose a novel generation method of visually protected images that can hold the luminance gradient direction of plain ones, referred to as gradient-preserving images. By using gradient-preserving images, HOG features can be extracted from the visually protected images. In addition, since gradient preserving images are irreversibly generated on the basis of a random number and a mathematical optimization, the protected images can be applied to various applications without any key management. In an experiment, an image recognition algorithm with a support-vector machine is carried out to confirm the effectiveness of the proposed method.

\section{Proposed Gradient-Preserving Image}
A generation method of gradient-preserving images and a HOG feature extraction from the generated images are explained here.

\subsection{Gradient-Preserving Images}
Let $x\in[0,1]^{H\times W}$ be a grayscale image with a size of $H\times W$.
Then, the luminance gradient direction map ${\rm GDM}(x)\in(-\pi/2,\pi/2)^{H\times W}$ of $x$ is defined as
\begin{align}
{\rm GDM}(x)_{h,w}={\rm arctan}\left(\frac{x_{h+1,w}-x_{h-1,w}}{x_{h,w+1}-x_{h,w-1}+\epsilon}\right) \quad , \label{GDM}
\end{align}
where $x_{h,w}$ and ${\rm GDM}(x)_{h,w}$ are scalar values at an index $(h,w), h=1,2,\ldots,H,\, w=1,2,\ldots,W$ in the spatial domain, respective, and $\epsilon$ is a small constant for avoiding division by a value of zero.

A gradient-preserving image $x^\prime\in[0,1]^{H\times W}$ should satisfiy
\begin{align}
{\rm GDM}(x)={\rm GDM}(x^\prime)\quad. \label{mokuhyou}
\end{align}

\subsection{Generation of Gradient-Preserving Images}
To generate a gradient-preserving image $x^\prime$ in accordance with Eq.(\ref{mokuhyou}), we define the mathematical optimization problem as
\begin{align}
{\rm argmin}_{x^\prime}&\,\, \|{\rm GDM}(x)-{\rm GDM}(x^\prime)\|_2 \quad, \label{saitekika1}\\
{\rm s.t.}&\,\, 0\leq x^\prime\leq 1 \quad . \nonumber
\end{align}
By approximating $x^\prime$ by using an array $s\in\mathbb{R}^{H\times W}$ and the sigmoid function ${\rm sigmoid}(\cdot)$ as
\begin{align}
x^\prime={\rm sigmoid}(s)\quad, \label{xdash s}
\end{align}
the optimization problem of Eq.(\ref{saitekika1}) can be transformed as
\begin{align}
{\rm argmin}_{s}\,\, \|{\rm GDM}(x)-{\rm GDM}({\rm sigmoid}(s))\|_2 \quad. \label{saitekika2}
\end{align}

Finally, a gradient-preserving image $x^\prime$ is obtained as $x^\prime={\rm sigmoid}(s^\prime)$, where $s^\prime$ is a local solution of Eq.(\ref{saitekika2}) solved by the steepest descent method.

\begin{figure}
 \begin{minipage}{0.47\hsize}
  \begin{center}
   \includegraphics[scale=0.2]{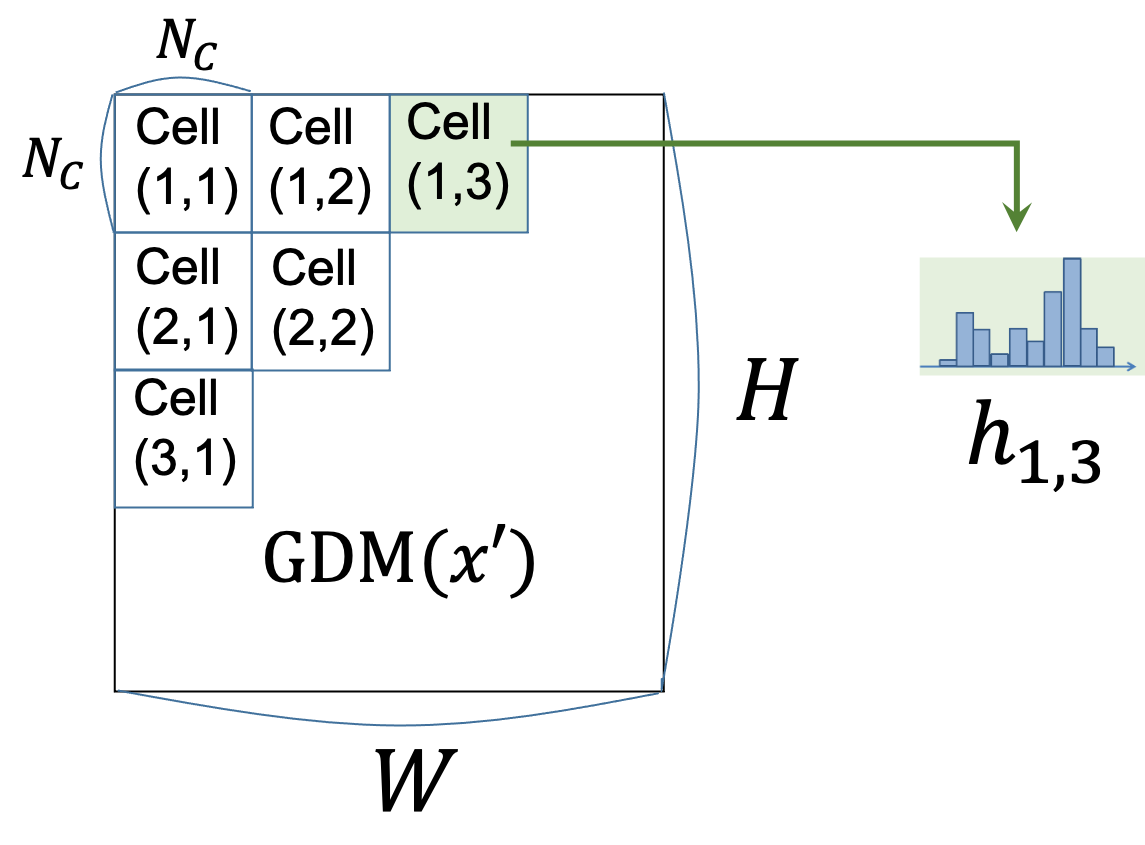}
   \caption{Cell, and histogram in each cell. \label{cell}}
  \end{center}
 \end{minipage}
 \hfill
 \begin{minipage}{0.47\hsize}
  \begin{center}
   \includegraphics[scale=0.2]{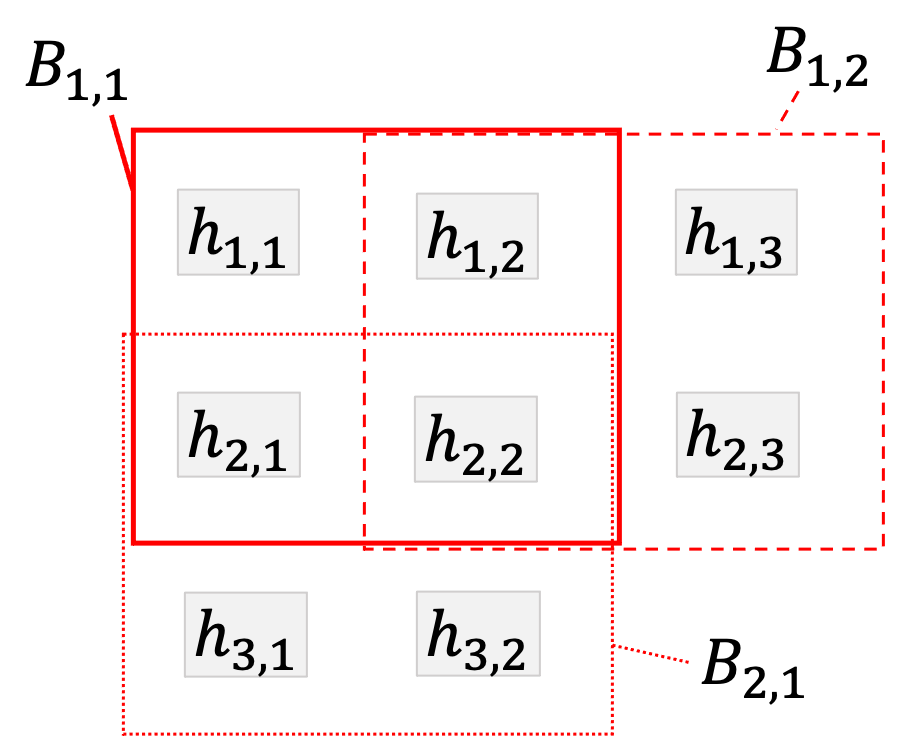}
   \caption{Relation between blocks and histograms $h_{i,j}$.  \label{block}}
  \end{center}
 \end{minipage}
\end{figure}

\subsection{HOG Feature Extraction}
HOG features can be extracted from gradient-preserving images $x^\prime$ as follows.

\noindent {\it step1}\,:\,\,Calculate luminance gradient direction map ${\rm GDM}(x^\prime)$ in accordance with Eq.(\ref{GDM}).

\noindent {\it step2}\,:\,\,Divide ${\rm GDM}(x^\prime)$ into non-overlapping blocks with a size of $N_c\times N_c$, called cells, and then obtain the histogram of ${\rm GDM}(x^\prime)$ for each cell, as $h_{i,j}\in\mathbb{R}^b,\,\,i\in\{1,2,\ldots,(H/N_C)\},j\in\{1,2,\ldots,(W/N_C)\}$ where $b$ is the number of bins in each histogram (see Fig.\ref{cell}).

\noindent {\it step3}\,:\,\,Define blocks by connecting multiple histograms as in Fig.\ref{block}, where each block consists of $2\times 2$ histograms in this paper, and histograms may be duplicated among blocks.
By using histograms in each block, block $B_{i,j}$ is given by
\begin{align}
B_{i,j}=h_{i,j}\oplus h_{i+1,j}\oplus h_{i,j+1}\oplus h_{i+1,j+1}\quad.
\end{align}

\noindent {\it step4}\,:\,\,Normalize elements in $B_{i,j}$ as below equation.
\begin{eqnarray}
 \label{normalize}
 \begin{array}{ll}
  B_{i,j}^{(n)}=B_{i,j}^{(n)}/\| B_{i,j}\|_2\,,&n=1,2,\ldots,4b\quad,\\
 \end{array}
\end{eqnarray}
where $B_{i,j}^{(n)}$ is the $n$-th element of $B_{i,j}$.

\noindent {\it step4}\,:\,\,Obtain the HOG feature vector of $x^\prime$ by concatenating all $\{B_{i,j}\}$.

The above procedure for extracting HOG features from a gradient-preserving image, is slightly different from that of the typical HOG feature extraction method\cite{HOG}, in which histograms are weighted with the magnitude of luminance gradients.
The proposed method focuses on preserving the luminance gradient direction of original images, so gradient-preserving images have no information on the magnitude of luminance gradients.

\section{Experiment}
To evaluate the effectiveness of the proposed method, we carried out an experiment with a face image dataset, called Extended Yale Face Database B\cite{Yale}.
This dataset contains 38 individuals, and 64 frontal facial images with 168$\times$192 pixels per each person.

\begin{figure}
 \begin{center}
  \includegraphics[scale=0.25]{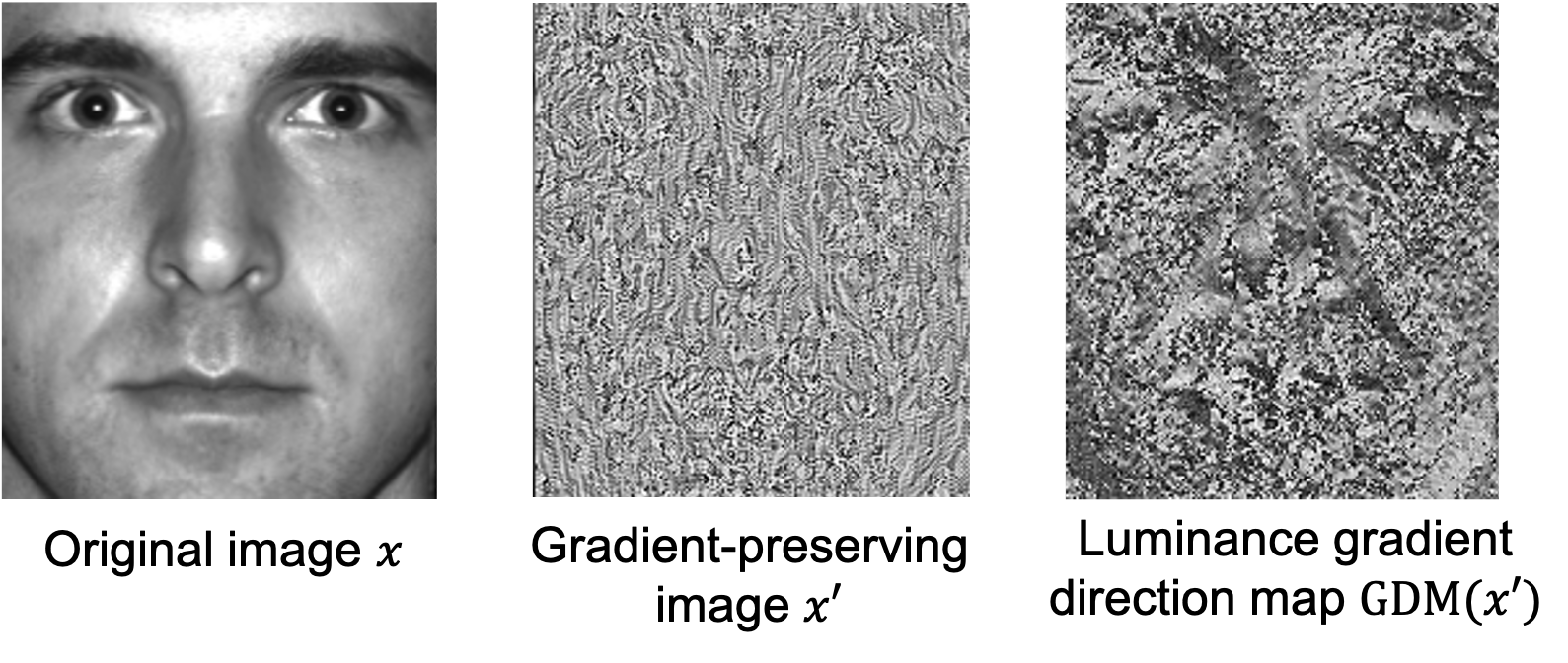}
  \caption{Example of original image $x$, gradient-preserving image $x^\prime$,  and luminance gradient direction map ${\rm GDM}(x^\prime)$. \label{imgs}}
 \end{center}
\end{figure}

\subsection{Evaluation of Visual Information}
Figure 3 shows an example of image x from the dataset, corresponding gradient-preserving image $x^\prime$, and the luminance gradient direction map ${\rm GDM}(x^\prime)$ of $x^\prime$.
As shown in the figure, $x^\prime$ had almost no visual information on $x$, so that it was difficult to identify a person from $x^\prime$.
Although ${\rm GDM}(x^\prime)$ slightly had visual information on a human face, it was still hard to identify who the original person is.

\begin{table}
\caption{Experimental result of face recognition using SVM \label{result}}
\begin{center}
\begin{tabular}{|l|l|}\hline
\quad\quad Feature extraction method&  \, Accuracy \\ \hline \hline
\begin{tabular}{l}{\bf Proposed (visually protected)}\end{tabular}&\begin{tabular}{l}0.9918\end{tabular} \\ \hline
\begin{tabular}{l}Conventional 1 (non-protected)\end{tabular}&\begin{tabular}{l}0.9942\end{tabular} \\ \hline
\begin{tabular}{l}Conventional 2 (non-protected)\end{tabular}&\begin{tabular}{l}0.9359\end{tabular} \\ \hline
\end{tabular}
\end{center}
\end{table}

\subsection{Face Recognition Experiment}
The dataset was randomly divided in half for training and testing.
By using the data, a face recognition experiment was carried out with a linear support vector machine (SVM). To extract feature vectors, the following three methods were applied.
\begin{itemize}
\item Proposed: Generate gradient-preserving images from the dataset and extract HOG features by using the proposed method.
\item Conventional 1: Directly extract HOG features from the dataset on the basis of a standard procedure\cite{HOG}.
\item Conventional 2: Extract features by applying Eigenface\cite{eigface} as a typical face recognition algorithm, to the dataset.
\end{itemize}
We used $N_c=8$ and $b=9$ as hyper parameters for the HOG feature extraction, and the dimension of feature vectors extracted for Eigenface was 150.

TABLE \ref{result} shows the experimental result.
From the table, the proposed method was confirmed to have almost the same accuracy as that of conventional method 1, which was better than conventional method 2, while protecting visual information on plain images.

\section{conclusion}
In this paper, we proposed a generation method of visually protected images, which allow us to directly extract HOG feature vectors.
The proposed method was demonstrated to be effective in a face recognition experiment with a SVM algorithm.

\bibliographystyle{IEEEbib}
\bibliography{ref_kitayama20190325}

\end{document}